\documentclass[conference]{IEEEtran}
\IEEEoverridecommandlockouts
\usepackage{cite}
\usepackage{amsmath,amssymb,amsfonts}
\usepackage{algorithmic}
\usepackage{graphicx}
\usepackage{textcomp}
\usepackage{xcolor}
\usepackage{gensymb}
\usepackage{dsfont}
\usepackage{url}
\usepackage{lipsum}
\usepackage{tabularx,booktabs}

\def\BibTeX{{\rm B\kern-.05em{\sc i\kern-.025em b}\kern-.08em
		T\kern-.1667em\lower.7ex\hbox{E}\kern-.125emX}}
\begin{document}
	
	\title{Large-Scale Cell-Level Quality of Service Estimation on 5G Networks Using Machine Learning Techniques\\
	}
	
	\author{\IEEEauthorblockN{M. Tu\u{g}berk \.{I}\c{s}yapar}
		\IEEEauthorblockA{
			\textit{Huawei Turkey R\&D Center}\\
			tugberk.isyapar1@huawei.com}
		\and
		\IEEEauthorblockN{Ufuk Uyan}
		\IEEEauthorblockA{
			\textit{Huawei Turkey R\&D Center}\\
			ufuk.uyan1@huawei.com}
		\and
		\IEEEauthorblockN{Mahiye Uluya\u{g}mur \"{O}zt\"{u}rk}
		\IEEEauthorblockA{
			\textit{Huawei Turkey R\&D Center}\\
			mahiye.uluyagmur.ozturk@huawei.com}
	}
	
	\maketitle
	
	\begin{abstract}
		This study presents a general machine learning framework to estimate the traffic-measurement-level experience rate at given throughput values in the form of a Key Performance Indicator for the cells on base stations across various cities, using busy-hour counter data, and several technical parameters together with the network topology. Relying on feature engineering techniques, scores of additional predictors are proposed to enhance the effects of raw correlated counter values over the corresponding targets, and to represent the underlying interactions among groups of cells within nearby spatial locations effectively. An end-to-end regression modeling is applied on the transformed data, with results presented on unseen cities of varying sizes.
	\end{abstract}
	
	
	\section{Introduction}
	Transition into the latest 5G technology brings in an immense increase in both traffic and speed demands together with evermore concurrently online users and machines, highlighting the role of cell performance evaluated by numerous \textit{Key Performance Indicators} (KPI's) derived from associated counter statistics. As opposed to previous telecommunication technologies, 5G requires KPI calculations more frequently by the network planning side, focusing on situation assessment and capacity expansion aspects to meet the strict requirements. 

	Traditional methods to obtain KPI values involve road tests that incur high costs in terms of labor, equipment, and time resources. However, counter values from thousands of cells are systematically recorded in databases alongside the historical KPI values. Using machine learning techniques, much needed functions that map collected data to KPI's can be approximated so as to be integrated within planning services.
	
	Assessing the experience rate solemnly based on traffic statistics is a challenging problem, since several factors such as the radio propagation environment, network coverage, traffic volume, number and distribution of users in different places, interference among groups of cells in nearby locations, and service model should be considered simultaneously. Machine learning models come handy in such difficult scenarios by revealing complex nonlinear relations within huge volumes of data in the form of generalizable knowledge patterns.
	
	Estimation of KPI's that focus on the throughput levels gained some interest in the literature. Obtaining and transforming counter values over cells, gradient-boosted trees are employed to predict throughput rates in \cite{minovski2021throughput}, with particular emphasis on signal quality related predictors. Over large-scale 4G networks, channel quality indicators together with user and traffic statistics from cell counters are utilized to estimate the downlink perceived throughput by employing deep neural networks in \cite{diouf2022finding}. 
	
	Similarly, for 4G networks, signal-related technical parameters in combination with downlink time measurements are used as inputs to machine learning models to classify whether throughput problems exist or not \cite{alho2021machine}. Context information related to the scenery around cells and associated distances are incorporated into modeling efforts in addition to the previously outlined variables for instantaneous throughput prediction \cite{samba2017instantaneous}. 
	
	Instead of tabular data regression and classification solutions covered so far, in \cite{schmid2019deep}, the authors propose to predict throughput using both historical data and influential factors within deep recurrent neural networks, without considering the spatial dependencies among cells. Rather than estimating the throughput directly, \textit{Signal-to-Interference-and-Noise-Ratio} is predicted using auto-regressive artificial neural networks in \cite{ullah2020machine} to increase the overall throughput of 5G networks remarkably. More recently, effective bandwidth prediction for 5G networks are formulated as a time series forecasting problem using actual data from driving scenarios \cite{lin2022bandwidth}, indicating the very dynamic nature of the mobile communication services.
	
	In this study, we have made the following contributions:
	
	\begin{itemize}
		\item Without resorting to time and signal quality related variables as inputs, we modeled the traffic measurement level satisfaction rate using cell counter values of mobile traffic statistics and technical parameters.
		\item We enriched the set of input features by deriving several predictors using the raw counter values.
		\item We incorporated spatial dependencies among nearby cells that might interfere with each other by defining a notion of neighborhood and the conditions for such interference together with a systematic feature engineering approach to enlarge the set of predictors influencing the KPI values.
		\item In this paper, we present an end-to-end machine learning study at a large geographical scale, with thousands of base station cells from several cities. Performance results are presented for real-case scenarios involving multiple urban settlements of varying sizes. To our knowledge, no previous studies that propose solutions to the problem in question exist at this scale.
	\end{itemize}	

	This study highlights a number of open research problems in subsequent sections, providing an elaborated description of machine learning modeling efforts, experimental evaluation procedures, and an overall review of important points together with directions for further research, in order.
	
	\section{Machine Learning Modeling}
	
	\subsection{Problem Definition}
	
	In our study, we intend to evaluate the mobile network performance at a large scale using a KPI formulation derived from throughput rates and associated downlink time readings, so that the quality of service can be expressed as a real number within the interval $[0, 1]$. A transition from values near $0$ towards $1$ implies poor to high-quality cell performance. The KPI in question, which will be referred to as the \textit{traffic-measurement-level experience rate at throughput-X} (TMLER-X), has the following formulation.
	
		\begin{equation}
		Y(X, i) = 1 - \dfrac {\sum_{i=1}^{X}\Phi^{DL}_i} {\sum_{i=1}^{B}\Phi^{DL}_i} \times \dfrac {\sum_{i=1}^{N-1}T^{DL}_i} {\sum_{i=1}^{N}T^{DL}_i}\label{srt_x}
		\end{equation}
	
	TMLER-X to assess the downlink traffic performance is introduced in \eqref{srt_x}, where $Y$ stands for the corresponding KPI value for a cell $i$ at a given throughput rate $X$. Hence, we obtain different indicator readings for varying throughput rates. Following the subtraction operation on the right-hand side of \eqref{srt_x}, the multiplicative term can be broken down into two parts in order. 
	
	The first multiplicand focuses on the user downlink \textit{throughput distribution} of the cell as downlink traffic volumes at different bins in the forms of increasing throughput rates in Mbps within the range $[0, \infty)$. Given the particular problem at hand, the number of bins is adopted as $B=15$, with a regular increase of $5$ Mbps until the throughput of $40$ Mbps is attained, following which the intra-bin difference values increase to $10$, $50$, $100$, $300$, and $500$ Mbps, respectively, while the last bin is particularly dedicated to any throughput rate exceeding $1000$ Mbps. The overall term is expressed as a ratio of the downlink volume up to and including the bin corresponding to the $X$ Mbps on the open end of the interval to the whole downlink volume. In other words, it will reach a maximum value of $1$ whenever all of the associated traffic is provided at a rate less than $X$ Mbps, and a minimum value of $0$ will be attained if all the traffic is realized at a rate greater than or equal to $X$ Mbps. The problem in question has $X$ value defined to be $100$ Mbps, which is a relatively high throughput rate, so that modeling gains a challenging aspect, and obtained TMLER-X values exhibit higher variance.
	
	The second multiplicand works on the downlink time readings per resource block in ms, and is expressed as the ratio of the total time spent in every slot other than the last one, in which the buffer becomes empty, to the overall time. The term reaches a value of $1$ whenever no time is spent on the last slot. Ideally, the two terms of the multiplication are highly correlated as larger traffic volumes should be associated with higher resource utilization in terms of the time component. Hence, whenever the service provides throughput rates no less than $X$ Mbps with good resource utilization, the multiplied component approaches a value of $0$, making the overall $Y$ value closer to $1$, indicating superior cell performance. Conversely, at a given $X$ Mbps, if the services are provided mostly at a lesser rate, or resource utilization is poor, the multiplicative term attains values closer to $1$, making TMLER-X approach $0$, implying inferior cell behavior. 
	
	TMLER-X makes the most sense when it is applied over the average statistics of counters collected during the same busy hour periods, in which the cell utilization is assumed to be the highest. Constituents of busy-hour traffic data acquired from multiple cells on base stations in several cities can be used in conjunction with the technical and spatial indicators per cell as predictors to approximate functions as models that yield the target KPI using machine learning techniques. As the corresponding KPI values are continuous, regression methods are utilized in the study. 
	
	In practice, network planning and simulation based exclusively on single-cell traffic counter values is a challenging task, due to the high variance such data exhibits innately. It is necessary to include cell-related technical parameters such as the frequency band and bandwidth that indicate the telecommunication generation and the service capacity as well as the cell coverage and topology, alongside with the characteristics of covered areas such as dense urban and rural regions, and the spatial dependencies among groups of associated cells, since all are highly influential factors over KPI readings.
	
	Incorporating technical parameters requires using machine learning models that are capable of handling both categorical and numerical features. Additional predictors are needed to be calculated to enhance predictive capabilities of the developed model by making use of domain knowledge expertise. So as to attain scene related information, additional indicators should be derived using descriptive map API's. Finally, a notion of neighborhood to identify interfering cells must be established using predetermined parameters within dedicated feature engineering techniques. All these approaches transform the input space of the problem, and our study primarily offers contributions on obtaining an enriched set of features, which will be elaborated in the following subsections.

	\subsection{Data Description and Processing}
	
	The raw data is collected from multiple undisclosed operators in several cities of distinct sizes. The running assumption is that the counter values are averaged during the same busy hour periods for each operator. So as to assess the performance better in unseen cases, $6$ cities of varying sizes out of nearly $40$ entities are strictly moved into the test set. Note that here the term city is used to refer to a single operator within an urban settlement, and some large cities indeed have more than operator. The data is private and contains hundreds of thousands of cells, however, we restrict the modeling efforts to the cells that have their physical resource block utilization of at least 20\%, so that it makes practical sense to include them within the network planning procedures. This way, we end up with more than 30000 and 10000 cells in the train and test sets, respectively. 
	
	\begin{table}[htbp]
		\caption{Cell Data Fields}
		\begin{center}
			\begin{tabular}{|c|c|c|}
				\hline
				\textbf{Predictor} & \textbf{Category}& \textbf{\textit{Description}} \\
				\hline
				Cell Name & Categorical & Unique cell identifier  \\
				\hline
				Longitude & Numeric & Positional identifier of cell \\
				\hline
				Latitude & Numeric & Positional identifier of cell \\
				\hline
				Azimuth & Numeric & Directional identifier of cell \\
				\hline
				DuplexMode & Categorical & TDD or FDD \\
				\hline
				DLNarfcn & Numeric & Downlink frequency encoding \\
				\hline
				FrequencyBand & Categorical & Generation and NR-band \\
				\hline
				DLBandwith & Numeric & Downlink bandwidth \\
				\hline
				TxRxMode & Categoric & Transmitter and receiver antennas\\
				\hline
				SubframeAssignment & Categoric & Subframe patterns\\
				\hline
				SpecialPatterns & Categoric & Special patterns for S blocks \\
				\hline
				DLPRBAvail & Numeric & Available PRB's for downlink \\
				\hline
				DLPRBUsage & Numeric & Downlink PRB usage as \% \\
				\hline
				ULPRBAvail & Numeric & Available PRB's for uplink \\
				\hline
				Height & Numeric & Height in meters\\
				\hline
				OnlineUserNumber & Numeric & Average number of online users \\
				\hline
				DLConcurrentUser & Numeric & Concurrent downlink users\\
				\hline
				ULConcurrentUser & Numeric & Concurrent uplink users\\
				\hline
				DLTraffic & Numeric & Downlink traffic volume in MB\\
				\hline
				ULTraffic & Numeric & Uplink traffic volume in MB\\
				\hline
				TotalTraffic & Numeric & Total traffic volume in MB\\
				\hline
			\end{tabular}
			\label{feat_descr}
		\end{center}
	\end{table}
	
	Raw data comprises of the columns listed in Table \ref{feat_descr} per cell, each of which is identified by a unique \textit{Cell Name} as the concatenation of its site id followed by its own id, and the operator. Listed set of features contain positional and directional information to locate cells together with the technical parameters that determine the role and capacity of the cell, such as the frequency band, bandwidth, and the number of antennas as well as their configurations. Next, the counter values that are obtained during the busy hours are listed. Since our task focuses on cell performance in the downlink traffic direction, we have the average allocated \textit{Physical Resource Block} (PRB) \textit{Usage} and the number of online users in conjunction with the concurrently online user statistics with associated traffic volumes, whose values are likewise provided for the uplink traffic. 
	
	As previously outlined, each cell is identified by a single record in the data files. However, additional features are needed to openly relate the existing counter values, and to model both the dependencies among associated cells and the scene information around them. 
	
	\begin{table}[htbp]
		\caption{Derived Cell Features}
		\begin{center}
			\begin{tabular}{|c|c|}
				\hline
				\textbf{Derived/Extracted Feature} & \textbf{\textit{Description}} \\
				\hline
				$Tx$ &  Number of transmitters  \\
				\hline
				$Rx$ &  Number of receivers  \\
				\hline
				$P^{DL}_u = P^{DL} * RP^{DL} / 100$ &  DL PRB's used \\
				\hline
				$U^{DL}_{C} / U^{UL}_{C}$ &  DL-to-UL Concurrent users \\
				\hline
				$UP^{DL}_{C} = U^{DL}_{C} / P^{DL}$ &  DL users to PRB's \\
				\hline
				$UP^{DL}_{C} = U^{DL}_{C} / P^{DL}_u$ &  DL users to used PRB's \\
				\hline
				$UP^{UL}_{C} = U^{UL}_{C} / P^{UL}$ &  UL users to PRB's \\
				\hline
				$VU^{DL}_{C} = V^{DL} / U^{DL}_{C}$ &  DL volume speed \\
				\hline
				$VU^{UL}_{C} = V^{UL} / U^{UL}_{C}$ &  UL volume speed \\
				\hline
				$V^{DL}_t = V^{DL} / Tx$ &  DL volume per unit \\
				\hline
				$VB^{DL} = V^{DL} / B^{DL}$ &  DL traffic to bandwidth \\
				\hline
				$VRP^{DL} = V^{DL} / RP^{DL}$ &  DL volume per 1\% PRB usage \\
				\hline
				$VUP^{DL}_{C} = V^{DL} / U^{DL}_{C} / P^{DL}$ &  DL traffic per co-user per PRB \\
				\hline
				$VUP^{DL}_{Cu} = V^{DL} / U^{DL}_{C} / P^{DL}_u$ &  DL traffic per co-user per used PRB \\
				\hline
				$VUP^{UL}_{C} = V^{UL} / U^{UL}_{C} / P^{UL}$ &  UL traffic per co-user per PRB \\
				\hline
			\end{tabular}
			\label{feat_derived_i}
		\end{center}
	\end{table}
	
	First set of derived features emerge as a result of the subsequent feature engineering efforts over raw measurements, and they are listed in Table \ref{feat_derived_i}. The majority of features is expressed as rates of traffic per resource and user, alongside with users per resource, and a further combination of the two. One thing to note is that the number of concurrent users in both downlink and uplink directions is intended to be used as substitutes of time-related counter values, which are not accessible to us.
	
	Second set of enriched features stem from the scenery information around each cell. Using open-source map API's, we calculated means and variances of location information at near, quasi-near, and far distances from every cell. The categories we have included are urban areas, regions with water such as rivers and seas, forestry and scrub areas together with lower plains, open areas, and others, which are assumed to have prominent effects on the cell coverage.
	
	Next, we need to include features derived from related cells for each cell. Primarily, we need to define the notion of \textit{neighborhood} of a cell, considering their positions on the grid as well as the direction towards which they are facing.	For the sake of simplicity, we assume that all the cells are placed in a two-dimensional flat grid with $0$ height.
	
	 \textit{Neighborhood} of a cell is adopted as a region bounded within a square, whose center is $500$ meters away from the cell's position towards the direction indicated by its azimuth. Once the center point is established, the four corners of the square will be $1200$ meters away from the center along azimuth values of $315 \degree$, $45 \degree$, $135 \degree$, and $225 \degree$ plus the cell azimuth in modulus $360 \degree$. Note that regardless of downlink PRB usage values, which we had previously restricted to be not less than $20 \%$ for modeling efforts, all the cells that fall within the computed square are selected as \textit{neighbors}. Furthermore, the operators of each neighbor cell must be the same in all cases.
	
	\begin{figure}[htbp]
		\centerline{\includegraphics[width=0.45\textwidth]{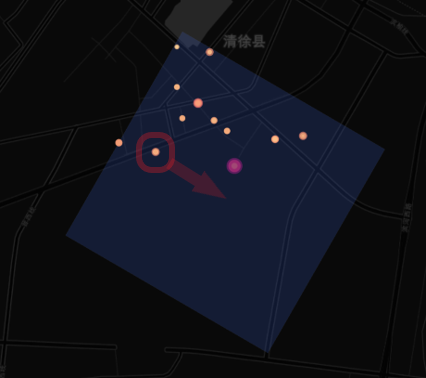}}
		\caption{Neighbors of an example cell within the bound box facing the direction illustrated with an arrow.}
		\label{neigh}
	\end{figure}
	
	Neighborhood determination procedure is illustrated in Fig.~\ref{neigh}. The cell for which the neighbors are determined resides within the bounding box, whose azimuth value of $120 \degree$ is depicted with an arrow. First, we go along the arrow for $500$ meters and reach the center. Next, the four corners are determined to form the square box, inside which only the neighbor cells reside, each of which is indicated with a dot that gets larger with a color shift towards purple as the TMLER-X value for $X=100$ Mbps gets diminished, implying lower satisfaction rates. Note that some of the cells that are behind the cell in question are also included within the neighborhood, since azimuth radiation patterns of a cell's electromagnetic signals are modeled to exhibit a limited coverage in the back areas \cite{niemela2005sensitivity}.
	
	Once the neighbors are established for each cell, the next step is to extract further features from neighbors. Within the scope of this study, we resorted to statistical feature engineering techniques in conjunction with features based on comparison. First, we narrow down the number of neighbors so that the feature extraction process is exclusively applied to the set of cells that might \textit{interfere} with the cell in question. The \textit{interfering} cells are defined as follows, which can be directly translated into an algorithm.
	
	\begin{itemize}
	
		\item Cells at the same location on the same site that have an absolute azimuth difference with the own cell less than or equal to $100 \degree$
		\item All cells excluding the ones on the same site, which are at a distance no higher than $150$ meters
		\item The cells that are at least $150$ and at most $1000$ meters apart, which fall within the field of view of the cell that is assumed to be $\pm 60 \degree$ of own azimuth, and that have the cell within their fields of view
		
	\end{itemize}

	So as to check whether two cells can actually see each other, azimuth value of the distance vector from the first cell to the latter together with the corresponding back azimuth value as the angle of the distance vector from the second cell to the first, which is a mere $180 \degree$ opposite of the angle of the distance vector, are compared against the limits of the fields of view of the corresponding cells. Each cell is assumed to have a coverage area of $120 \degree$ that is symmetric around its azimuth axis for practical purposes.
	
	Now that the interfering cells have been determined, statistical feature extraction over associated predictors can be carried out. Since several statistics of each raw and derived feature can be subject to this procedure, we would end up with an enormous set if we took the exhaustive approach. Instead, based on trial and error via initial modeling attempts on the data, we decided to utilize the following list of features.
	
	\begin{itemize}
		
		\item Height 
		\item Online User Number
		\item Available Downlink PRB's
		\item Downlink PRB Usage
		\item Used Downlink PRB's
		\item Downlink Traffic
		\item Uplink Traffic
		\item Downlink Concurrent Users
		\item Uplink Concurrent Users
		\item Rate of Downlink Traffic to Downlink Concurrent Users
		\item Rate of Uplink Traffic to Uplink Concurrent Users
		\item Downlink Traffic Per Downlink Concurrent User Per Available PRB

	\end{itemize}
	
	Regarding each feature listed above, following statistics are extracted as interfering cell features.
	
	\begin{itemize}
		\item Ratio of cells that have their corresponding feature values less than or equal to the cell's value
		\item Ratio of cells that have their corresponding feature values greater than the cell's value
		\item Mean, mean plus standard deviation, mean minus standard deviation, median, max, $25^{th}$ and $75^{th}$ percentiles of all interfering cells' feature values 
		\item Normalized counts of features of interfering cells that reside on bins with endpoints populated from the ordered set $\{x^{th} \text{percentile } |\, x \in \{0, 25, 50, 75, 100 \}\}$.
	\end{itemize}

	In order to effectively discriminate the extracted features above from own values, additional indicators based on comparison techniques are calculated. To compare two values, the following operations are carried out, and the resultant values are added to the set of extracted features.
	
	\begin{itemize}
		\item $1$ if the difference between own value and interfering cell feature is positive, $0$ otherwise
		\item \textit{Relative accuracy} \cite{Tofallis_2015} as $(\ln f_n - \ln f_c)$ where $f_n$ is the interfering cell feature and $f_c$ is the own value
		\item Comparison of the magnitude of difference to own value as in $( \ln |f_c - f_n| - \ln f_c)$
	\end{itemize}

	Feature extraction procedure is completed after a few additional indicators such as the number of neighbors and interfering cells, as well as statistics of the distance values between related cells and own cell in corresponding contexts are calculated. At the end, the feature set size swells up to almost $300$. 
	
	The set of predictors and the target KPI values are subject to general data preprocessing techniques including but not limited to reflecting over density-axis to make a left-skewed distribution right-skewed, power transformations, scaling and standardization, depending on the performance of associated machine learning models. 

	\subsection{Regression Model}
	
	Data prepared in the previous section is devoid of an explicit temporal component, as all counter values per operator are collected at the same time and there are single entries per cell. Furthermore, the features we have extracted are put forward to model the spatial correlations. At this point, a flexible machine learning model that is capable of capturing and generalizing nonlinear relations over raw and derived features is needed. For this purpose, CatBoost \cite{DBLP:journals/corr/abs-1810-11363} gradient-boosting algorithm is adopted, due to its versatile performance that is not wholly dependent on selected hyperparameters, scalability over big data sets, and innate support for categorical features.

	\section{Experimental Evaluation}
	
	In this section, a number of experiments will be carried out to assess the generalization performance of our model on the test set. First, hyperparameter search and optimization using both cross-fold validation on train data and validation through test cities will be described. 
	
		\begin{table}[htbp]
		\caption{Hyperparameter Search}
		\begin{center}
			\begin{tabular}{|c|c|}
				\hline
				\textbf{Parameter} & \textbf{\textit{Search Values}} \\
				\hline
				\texttt{n\_estimators} &   $[250, 3000]$$^{\mathrm{a}}$ or $4000$$^{\mathrm{b}}$\\
				\hline
				\texttt{early\_stopping\_rounds} & None$^{\mathrm{a}}$ or 500$^{\mathrm{b}}$\\
				\hline
				\texttt{learning\_rate} & $[10^{-4}, 1]$$^{\mathrm{c}}$\\
				\hline
				\texttt{l2\_leaf\_reg} & $[0, 10]$\\
				\hline
				\texttt{random\_strength} & $[0, 5]$\\
				\hline
				\texttt{colsample\_bylevel} & $[0.01, 1]$$^{\mathrm{c}}$\\
				\hline
				\texttt{boosting\_type} & $\{\mathtt{Ordered}, \mathtt{Plain}\}$\\
				\hline
				\texttt{bootstrap\_type} & $\{\mathtt{Bayesian}, \mathtt{Bernoulli}, \mathtt{MVS}\}$\\
				\hline
				\texttt{objective} & $\{\mathtt{RMSE}, \mathtt{Huber:delta=1.0}\}$\\
				\hline
				\texttt{use\_best\_model} &   \texttt{False}$^{\mathrm{a}}$ or \texttt{True}$^{\mathrm{b}}$\\
				\hline
				\texttt{bagging\_temperature} & $[0, 10]$$^{\mathrm{d}}$\\
				\hline
				\texttt{subsample} & $[0.1, 1]$$^{\mathrm{c, e}}$\\
				\hline
				\texttt{grow\_policy} & $\{\mathtt{SymmetricTree}, \mathtt{Lossguide}\}$$^{\mathrm{f}}$\\
				\hline
				\multicolumn{2}{l}{$^{\mathrm{a}}$Cross-fold validation setting.}\\
				\multicolumn{2}{l}{$^{\mathrm{b}}$Validation on test data setting.}\\
				\multicolumn{2}{l}{$^{\mathrm{c}}$Search in logarithmic scale.}\\
				\multicolumn{2}{l}{$^{\mathrm{d}}$Only if bootstrap type is \texttt{Bayesian}.}\\
				\multicolumn{2}{l}{$^{\mathrm{e}}$Only if bootstrap type is \texttt{Bernoulli} or \texttt{MVS}.}\\
				\multicolumn{2}{l}{$^{\mathrm{f}}$Only if boosting type is \texttt{Plain}.}\\
			\end{tabular}
			\label{hyperparameters}
		\end{center}
	\end{table}

	Even though default parameters of CatBoost do not require a lot of tweaking to end up with a reasonable model, the algorithm still comes with a large number of variables that change the way the ensemble is built. The complete hyperparameter list as input to the optimization procedure is summarized in Table \ref{hyperparameters}. Optuna Hyperparameter Optimization Framework \cite{DBLP:journals/corr/abs-1907-10902} is used to reveal the best parameters. For reproducibility purposes, all seed values are adopted as $0$. 
	
	Next, the evaluation metrics will be defined, and the experimental settings involving cross-fold validation and the sample weighing over the optimum models will be described thoroughly together with the incurred results.
	
	\begin{table*}
		\caption{Performance Through Optimum Hyperparameters}
		\label{eval_results}
		\begin{center}
			\begin{tabularx}{0.62\textwidth}{@{} l *{8}{c} c @{}}
				\toprule
				& & \multicolumn{2}{c}{\textit{Cross-fold Validation}} & \multicolumn{2}{c}{\textit{Test as Validation}} & 
				\multicolumn{2}{c}{\textit{Test as Validation}}  \\
				& & \multicolumn{2}{c}{\textit{(RMSE)}} & \multicolumn{2}{c}{\textit{(RMSE)}} & 
				\multicolumn{2}{c}{\textit{(Huber, one-minus)}}  \\
				\addlinespace
				City-Operators
				& Instances & MAPE & $P_6$ & MAPE & $P_6$ 
				& MAPE & $P_6$ \\ 
				\midrule
				
				Combined Train    & 35856    & 2.73  & 90.36\%    & \textbf{2.43}    & \textbf{93.04}\%    & 2.45  & \textbf{93.04}\%  \\
				Combined Test     & 12255    & \textbf{2.74}  & 92.04\%    & 2.75    & 92.04\%    & 2.75  & \textbf{92.14}\%  \\ 
				
				\addlinespace
				
				City-Operator A   & 5289     & 2.25  & 97.13\%    & \textbf{2.24}    & \textbf{97.22}\%   & 2.26  & 97.13\%  \\ 
				City-Operator B   & 826      & \textbf{3.00}  & \textbf{90.31}\%    & 3.01    & 89.23\%   & 3.09  & 89.83\%  \\ 
				City-Operator C   & 1013     & \textbf{2.81}  & 88.55\%    & 2.83    & \textbf{89.44}\%   & 2.83  & 88.65\%  \\ 
				City-Operator D   & 4044     & 3.38  & 86.62\%    & 3.40    & 86.60\%   & \textbf{3.36}  & \textbf{87.19}\%  \\ 
				City-Operator E   & 275      & \textbf{2.53}  & \textbf{93.09}\%    & 2.57    & 92.00\%   & 2.60  & 91.27\%  \\ 
				City-Operator F   & 808      & 2.52  & \textbf{91.58}\%    & \textbf{2.46}    & 91.46\%   & 2.51  & 91.34\%  \\ 
				\bottomrule
			\end{tabularx}
		\end{center}
	\end{table*}

	\subsection{Evaluation Metrics}
	
	As target TMLER-X values are within a range of $[0, 1]$, they are treated as percentage values by multiplying each with a factor of $100$. Next, a \textit{mean absolute percentage error} (MAPE) metric is defined in the following.
	
	\begin{equation}
		\text{MAPE}(Y,\, \hat{Y}) = \dfrac 1 N \sum_{i=1}^N |Y_i - \hat{Y}_i| \label{mape}
	\end{equation}

	In \eqref{mape}, $Y$ is the collection of ground truth, while $\hat{Y}$ is the corresponding prediction vector for $N$ instances of a given data set. Note that both $Y$ and $\hat{Y}$ are raw values of the targets multiplied by $100$. Moreover, a performance target assessment criterion is defined as follows.
	
	\begin{equation}
		\text{P}_{\theta}(Y,\, \hat{Y}) =  \dfrac {100} N \sum_{i=1}^N \mathds{1}_{\theta}(|Y_i - \hat{Y}_i| \leq \theta) \label{percent}
	\end{equation}
	
	\textit{Performance coverage}, $P_{\theta}$ in \eqref{percent}, determines the percentage of all cells, for which the absolute percentage error is less than or equal to a given a threshold $\theta$ within $[0, 100]$. Our study aims to achieve a coverage of at least $90\%$ of cells in the test set at $\theta = 6\%$. 

	\subsection{Cross-Fold Validation}
	In the first experimental setting, repetitive 7-fold validation over train set via stratified sampling on discretized bins of target values populated from the ordered set $\{0, 0.5, 0.7, 0.8, 0.9, 1.0\}$ is performed. Following hyperparameters yield the best results. 
	
	\begin{itemize}
		\item \texttt{boosting\_type}: \texttt{Ordered}
		\item \texttt{bootstrap\_type}: \texttt{MVS}
		\item \texttt{colsample\_bylevel}: $0.901$
		\item \texttt{depth}: $9$
		\item \texttt{l2\_leaf\_reg}: $8.731$
		\item \texttt{learning\_rate}: $0.038$
		\item \texttt{n\_estimators}: $2351$
		\item \texttt{objective}: \texttt{RMSE}
		\item \texttt{random\_strength}: $1.030$
		\item \texttt{subsample}: $0.977$
	\end{itemize}
	
	Once the optimum hyperparameters are determined, another CatBoost instance is fit using them on the whole train data. Train and test performance of the corresponding model are summarized in Table \ref{eval_results}. Thanks to the cross-fold validation, train and test performances are quite comparable in general, with no apparent overfitting or underfitting issues. 
	
	However, when we analyze the city-operator-wise performance, we see that regarding the entities $D$ and $B$, the error rate is much higher, while at the same time, predictions for the entities $A$, $E$, and $F$ achieve better-than-average performance. This suggests that the underlying dynamics in each city-operator exhibit high-variance, and particular data shift and mismatch issues \cite{10.5555/1462129} are encountered in the adopted splits. 
	
		\begin{table*}[htbp]
		\caption{Label Distribution Smoothing and Instance Weighing}
		\label{eval_results_lds}
		\begin{center}
			\begin{tabularx}{0.67\textwidth}{@{} l *{8}{c} c @{}}
				\toprule
				& & 
				\multicolumn{2}{c}{\textit{Test as Validation}} & 
				\multicolumn{2}{c}{\textit{Test as Validation}} &
				\multicolumn{2}{c}{\textit{Test as Validation}}  \\
				& &  \multicolumn{2}{c}{\textit{(RMSE, HO)}} & 
				\multicolumn{2}{c}{\textit{(RMSE, LDS)}}  &
				\multicolumn{2}{c}{\textit{(RMSE, LDS, HO)}}\\
				\addlinespace
				City-Operators
				& Instances & MAPE & $P_6$ & MAPE & $P_6$ 
				& MAPE & $P_6$ \\ 
				\midrule
				Combined Train [0.0, 1.0]  & 35856   & \textbf{2.43}  & 93.04\%    & 2.52    & 92.77\%    & \textbf{2.43}  & \textbf{93.70}\%  \\
				Combined Train (0.9, 1.0]  & 8350    & \textbf{1.73}  & \textbf{96.95}\%    & 2.01    & 96.25\%    & 2.00  & 96.08\%  \\
				Combined Train (0.8, 0.9]  & 12092   & \textbf{1.97}  & \textbf{97.47}\%    & 2.47    & 94.20\%    & 2.52  & 93.12\%  \\
				Combined Train (0.7, 0.8]  & 4750    & 2.92  & 89.01\%    & \textbf{2.88}    & 89.62\%    & 2.92  & \textbf{89.90}\%  \\
				Combined Train (0.5, 0.7]  & 4277    & 3.55  & 82.63\%    & 3.03    & 88.08\%    & \textbf{2.60}  & \textbf{93.55}\%  \\
				Combined Train [0.0, 0.5]  & 6387    & 3.09  & 87.26\%    & 2.68    & 90.73\%    & \textbf{2.33}  & \textbf{94.60}\%  \\
				
				\addlinespace
				
				Combined Test [0.0, 1.0]  & 12255   & \textbf{2.75}  & \textbf{92.04}\%    & 2.93    & 90.75\%    & 3.00  & 89.60\%  \\
				Combined Test (0.9, 1.0]  & 2782    & \textbf{2.38}  & \textbf{96.12}\%    & 2.65    & 94.21\%    & 2.54  & 93.85\%  \\
				Combined Test (0.8, 0.9]  & 5671    & \textbf{2.36}  & \textbf{95.87}\%    & 2.73    & 93.04\%    & 2.78  & 92.37\%  \\
				Combined Test (0.7, 0.8]  & 2417    & 2.93  & 89.74\%    & \textbf{2.86}    & \textbf{91.02}\%    & 3.08  & 88.08\%  \\
				Combined Test (0.5, 0.7]  & 1009    & 4.50  & 74.53\%    & \textbf{4.12}    & \textbf{76.91}\%    & 4.32  & \textbf{76.91}\%  \\
				Combined Test [0.0, 0.5]  & 376     & 5.38  & 65.69\%    & \textbf{5.25}    & \textbf{66.22}\%    & 5.84  & 60.11\%  \\
				\bottomrule
			\end{tabularx}
		\end{center}
	\end{table*}

	\subsection{Using Test Cities for Validation}
	Using test set instances as a validation set to optimize the training hyperparameters is the second approach we have taken. Here, two different sets of optimum parameters are determined. The first set of hyperparameters discovered using the standard \textit{root-mean-squared-error} (RMSE) as the loss criterion are listed below.
	
	\begin{itemize}
		\item \texttt{boosting\_type}: \texttt{Ordered}
		\item \texttt{bootstrap\_type}: \texttt{MVS}
		\item \texttt{colsample\_bylevel}: $0.901$
		\item \texttt{depth}: $8$
		\item \texttt{l2\_leaf\_reg}: $6.028$
		\item \texttt{learning\_rate}: $0.016$
		\item \texttt{n\_estimators}: $2987$
		\item \texttt{objective}: \texttt{RMSE}
		\item \texttt{random\_strength}: $2.725$
		\item \texttt{subsample}: $1$
	\end{itemize}
	
	The second hyperparameter search is based on the utilization of Huber loss with $\delta = 1$, and the target is transformed by subtracting all values from $1$ so that it has a right-skewed distribution. Corresponding set of optimum hyperparameters are listed subsequently.
	
	\begin{itemize}
		\item \texttt{boosting\_type}: \texttt{Plain}
		\item \texttt{bootstrap\_type}: \texttt{MVS}
		\item \texttt{grow\_policy}: \texttt{Lossguide}
		\item \texttt{colsample\_bylevel}: $0.729$
		\item \texttt{depth}: $7$
		\item \texttt{l2\_leaf\_reg}: $8.883$
		\item \texttt{learning\_rate}: $0.054$
		\item \texttt{n\_estimators}: $746$
		\item \texttt{objective}: \texttt{Huber:delta=1.0}
		\item \texttt{random\_strength}: $0.943$
		\item \texttt{subsample}: $0.343$
	\end{itemize}

	Evaluation results are listed in Table \ref{eval_results} along with the cross-fold validation. In general, tendency to overfit to the train data has become more prominent in this experimental setting, while the test performance remains almost unchanged. Even though city-operator-based metric readings exhibit some variance, the most general results outlined in the previous section hold true for the case of using the test set as a validation set during hyperparameter optimization. However, the second set of hyperparameters in this section truly indicate that simpler CatBoost models can achieve good performance with the appropriate set of parameters.
	
	\subsection{Sample Weighing on Optimum Model}
	Deeper analysis into the first set of hyperparameters in using the test set for validation purposes over discretized bins of the target values reveals that the performance abruptly changes depending on how densely the corresponding bins are distributed, as summarized in the first block columns of Table \ref{eval_results_lds}. Given that the performance deteriorates in sparser regions for both train and test sets, using \textit{Label Distribution Smoothing} (LDS) techniques for regression \cite{yang2021delving} to address the related issues is the reason this experiment is carried out.
	
	Using the aforementioned hyperparameters exactly as is, with only applying the LDS over $700$ equidistant bins via a Gaussian kernel of length $7$ and a standard deviation of $2$, training errors in denser regions increase, while the performance improves on the sparser bins, as summarized in the second block columns of Table \ref{eval_results_lds}. The same tendency is observed in test performance as well. However, both overall train and test errors have increased, and the issue stemming from the data set shifts between sparser regions of the train and test sets remain unaddressed.
	
	When we incorporate the LDS within hyperparameter search itself, the variance of incurred error among bins of the training set clearly diminishes, as the last block columns of Table \ref{eval_results_lds} indicate. Nonetheless, test performance significantly deteriorates as well, highly likely due to an overemphasized data shift in the sparser outlier-bearing regions when we try to replicate the train predictions over the test set.
	
	\section{Conclusion}
	In this study, we have presented a general machine learning methodology with an emphasis on feature design to utilize the initially existing cell-based predictors more effectively, and to represent the interactions between cells with higher proficiency, in order to provide a data-centric solution to predicting the service quality of cells in recently established 5G networks. 
	
	Scenery information around base stations is highly impactful in determining the data dynamics, as we have already used statistical map features. However, doing so, only part of the reality has been incorporated into modeling efforts. Utilizing powerful machine learning models to predict the scene information of many more points around the cell's location \cite{https://doi.org/10.48550/arxiv.2212.02130} will likely boost the proposed models henceforward. Similarly, interpolating the existing counter or derived features to discrete grid structures using realistic signal propagation models that take the physical nature of the issue, surrounding land forms and all kinds of constructs in three dimensions, together with human and equipment movements into consideration poses as an open research topic that will likely gain more interest in foreseeable future. 
	
	We eliminated the temporal dimension by restricting the obtained measurements to historical busy-hour averages, so that the status of cells can be assessed within the network planning procedures, rather than focusing on the innate spontaneous changes. Still, as mobile network capacity and traffic statistics are more dynamic than ever with the introduction of 5G services \cite{https://doi.org/10.48550/arxiv.2212.10869}, KPI values do inherit the tendency to change rapidly. We have shown that city and operator characteristics also exhibit considerable variances. Consequently, the framework we proposed will require retraining to capture the changing data characteristics. Last but not least, we further intend to analyze and identify the conditions in which retraining will be required through the lens of numerous transfer learning scenarios.

\bibliographystyle{IEEEtran}


	
\end{document}